\documentclass[letterpaper, 10 pt, conference]{ieeeconf}  

\IEEEoverridecommandlockouts                              

\usepackage{graphicx}
\usepackage{multirow}
\usepackage{amsmath}
\usepackage{booktabs}
\usepackage{comment}
\usepackage{flushend}
\usepackage{adjustbox}

\makeatletter
\let\NAT@parse\undefined
\makeatother

\usepackage[breaklinks,colorlinks]{hyperref}

\usepackage[capitalize]{cleveref}
\crefname{section}{Section}{Sections}
\Crefname{section}{Section}{Sections}
\crefname{table}{Table}{Tables
\Crefname{table}{Table}{Tables}}
\crefname{figure}{Fig.}{Figs.}
\Crefname{figure}{Figure}{Figures}

\title{\LARGE \bf
Advancements in Radar Odometry
}

\author{\authorblockN{Matteo Frosi, Mirko Usuelli, and Matteo Matteucci}
\authorblockA{
Department of Electronics Information and Bioengineering\\
Politecnico di Milano\\
Milan, Italy\\
{\tt\footnotesize \{name.surname\}@polimi.it}
}
}

\begin{document}

\maketitle
\thispagestyle{empty}
\pagestyle{empty}

\begin{abstract}
Radar odometry estimation has emerged as a critical technique in the field of autonomous navigation, providing robust and reliable motion estimation under various environmental conditions. Despite its potential, the complex nature of radar signals and the inherent challenges associated with processing these signals have limited the widespread adoption of this technology. 
This paper aims to address these challenges and simultaneously present an understanding about the current advancements in radar odometry estimation. First, we propose novel improvements to an existing state-of-the-art method, which are designed to enhance accuracy and reliability in diverse scenarios. Our pipeline consists of filtering, motion compensation, oriented surface points computation, smoothing, one-to-many radar scan registration, and pose refinement. In particular, we enforce local understanding of a scene by including additional information through smoothing (Gaussian kernels) and alignment (ICP), introduced by us in the existing pipeline. Then, we present an in-depth investigation of the contribution of each improvement to the localization accuracy. Lastly, we benchmark our system and state-of-the-art methods on all sequences of well-known datasets for radar understanding, i.e., the Oxford Radar RobotCar, MulRan, and Boreas datasets. In particular, Boreas includes scenarios with challenging weather conditions, such as snow or overcast, and, to our knowledge, it has never been used for evaluation or benchmarking in the literature. The proposed improvements allow to achieve superior results, on all scenarios considered and under harsh environmental constraints. 

\end{abstract}


\section{Introduction}\label{section:introduction}
Robust localization is necessary for numerous tasks in the real world, from autonomous driving and transportation, where continuous autonomy is required, to industrial robots, which usually operate in harsh environmental settings within underground mines, construction sites, or even in adverse weather conditions. Indeed, accurately estimating the odometry of a robot in large-scale scenarios is still a considerable challenge for autonomous navigation and all its applications.

Nowadays, state-of-the-art systems exploit exteroceptive sensors like cameras or LiDARs to perform accurate localization~\cite{nahavandi2022comprehensive}. Although the available systems reached a mature state, these sensors are sensitive to environmental artifacts, such as dust, rain, fog, and snow, but also to illumination and viewpoint changes~\cite{garg2004detection, charron2018noising, zhang2018robust}. Due to their longer wavelength, radar sensors prove to be largely resilient to these conditions~\cite{burnett2022we}. Moreover, they can penetrate certain materials, extending their detection area, and going further than the line of sight of LiDARs. Lastly, radars have been receiving increasing attention for various perception tasks, such as place recognition and loop closure detection~\cite{usuelli2023radarlcd}.

The most widely used type of sensor to achieve localization is the spinning radar, as it is able to provide accurate and dense data, in the form of polar images, representing the power returns at different range and azimuth values. One challenge of spinning radars is that the obtained measurements are inherently hard to interpret, due to the noise characteristics of the sensor~\cite{venon2022millimeter}, which lead to specific artifacts in the images, e.g., speckle noise, multi-path reflections, or ghost objects. The majority of localization systems in the literature extract hand-crafted features from intensity peaks in the polar images, perform robust data association, and compensate for motion distortion~\cite{cen2018precise, cen2019radar, hong2022radarslam, adolfsson2023cfear3}. Although state-of-the-art methods integrate local geometry into cost functions and exploit multiple radar sweeps for online incremental odometry, limited attention has been put into possible refinement and smoothing techniques, which further boost localization by using additional spatial information.

\begin{figure}[t]
    \vspace{.1em} 
    \centering
    \includegraphics[width=0.45\textwidth]{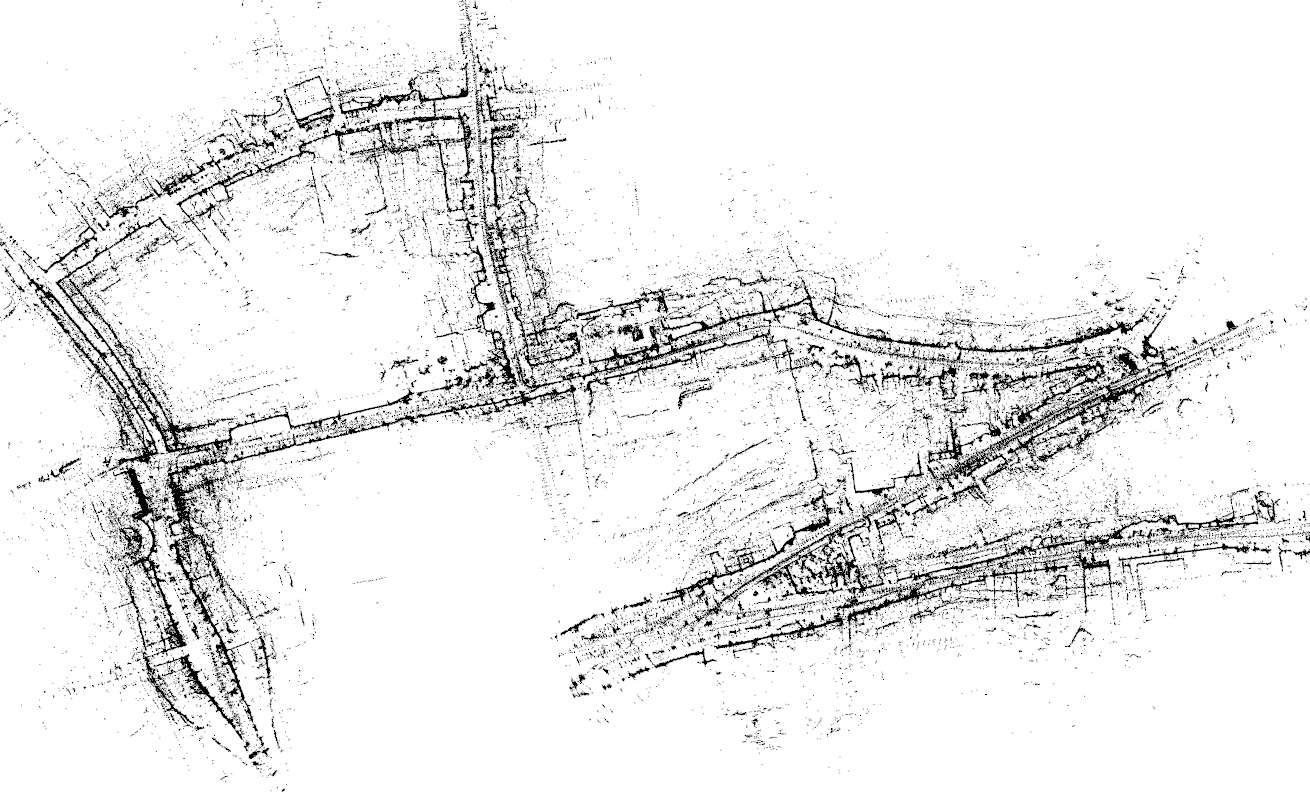}
    \caption{Map obtained by testing the proposed method on partial sequence 10-14-35 of the Oxford Radar RobotCar dataset~\cite{barnes2020oxford}.}
    \label{figure:map_reconstruction}
\end{figure}

This manuscript builds upon the \textit{Conservative Filtering for Efficient and Accurate Radar odometry} work, also known as CFEAR-3 Radar odometry~\cite{adolfsson2023cfear3}. The authors proposed (i) an efficient method for computing a sparse set of oriented surface points from the intensity peaks of a radar image, by applying a conservative filter and then analyzing local geometries. Then, they included (ii) an incremental scan matching approach that registers the latest scan to multiple non-consecutive scans jointly, by performing weighted minimization of point-to-point correspondence metric. With our work, we include several improvements and a new version of CFEAR that increases its accuracy, as it can be seen in Fig.~\ref{figure:map_reconstruction}, representing the map obtained on partial sequence 10-14-35 of the Oxford Radar RobotCar dataset~\cite{barnes2020oxford}.

We additionally include an extensive evaluation of the proposed method, along with multiple state-of-the-art systems for radar odometry estimation, on the most well-known datasets for radar data, i.e., the Oxford Radar RobotCar~\cite{barnes2020oxford}, MulRan~\cite{kim2020mulran}, and the recent Boreas~\cite{burnett_ijrr23} datasets. To our knowledge, we represent the first work on radar odometry in which Boreas has been used for benchmarking.

Summing up, we present multiple novel contributions.
\begin{itemize}
    \item A full radar odometry estimation pipeline consisting of filtering, motion compensation, oriented surface points computation, smoothing, scan-to-multi-keyframe registration, and pose refinement.
    \item An improved approach for computing the set of oriented surface points, as in (i), which consists of two optional and mutually exclusive methods for kernel smoothing.
    \item A pose refinement strategy, following (ii), which exploits the previously computed oriented surface points to perform direct point cloud alignment between temporally consecutive radar scans (not keyframes).
    \item An extensive evaluation of the proposed system and state-of-the-art methods on all sequences of the Oxford Radar RobotCar~\cite{barnes2020oxford}, MulRan~\cite{kim2020mulran}, and Boreas~\cite{burnett_ijrr23} datasets, considering that the last one has not been yet tested in the literature despite the variety of weather conditions recorded, as we show in the results.
    \item Finally, we make our C++ implementation available to the community, completely ROS-independent: \url{https://github.com/AIRLab-POLIMI/CFEAR4}.
\end{itemize}

\section{Related works}\label{section:related_works}
Due to the advantages of radars over other sensors, over the course of the last decade, many odometry estimation and mapping methods relying on radars have been proposed. 

A common practice is to utilize multiple automotive radar sensors~\cite{caesar2020nuscenes}, which currently offer range and azimuth resolutions similar to laser rangefinders, for estimating numerous relative target velocities (Doppler effect) that can assist in determining the robot’s motion~\cite{kellner2013instantaneous}. However, this method may not be the most effective for both odometry estimation and mapping due to the relatively low accuracy of velocity measurements compared to other sensors.

Attention then turned to the potential of harnessing the fundamental signal data from radars. From this data, prominent features can be extracted in a manner akin to vision-based systems. However, the process of extracting keypoints from radar data and utilizing them for direct data association has proven to be a complex task due to challenging modeling. 

Jose and Adams~\cite{jose2004relative} were the first to research the application of radars in outdoor localization and mapping. In particular, they proposed a feature detector that estimates the probability of target presence while augmenting their formulation to include radar cross-section as a discriminating feature. The works in~\cite{rouveure2009high} directly found the transformation between pairs of dense scans, using 3D cross-correlation.

Cen and Newmann~\cite{cen2018precise} then presented a method to extract stable keypoints in radar images, which are then used to perform scan matching, and accurately estimate the motion of the robot. The same authors later proposed an update~\cite{cen2019radar} to their radar odometry pipeline, which improved keypoint detection, and proposed a novel graph-matching strategy. 

Recently, studies that solely utilize radars have primarily concentrated on enhancing aspects of odometry~\cite{kung2021normal, aldera2019could, park2020pharao}. Noteworthy research on effective methods for odometry estimation is presented in the works of Burnett et al.~\cite{burnett2021we} and Park et al.~\cite{park2020pharao}. In the first approach, also referred to as YETI, the researchers measured the significance of motion distortion and demonstrated that Doppler effects need to be eliminated during both localization and mapping. In the second approach, known as PhaRaO, a direct method was suggested as an alternative to using feature-based methods. 

Adolfsson et al. proposed CFEAR-1~\cite{adolfsson2021cfear1} as an accurate radar odometry estimation method. First, a radar filter keeps only the strongest reflections per-azimuth that exceeds a threshold noise level. Then, the filtered radar data is used to incrementally estimate odometry by registering the current scan with a nearby keyframe. Registration is achieved through the modeling of local surfaces in the data and the minimization of a point-to-line metric. CFEAR-2~\cite{adolfsson2021cfear2} improved over the previous method in two ways. Drift is reduced by jointly registering the latest radar scan to a history of keyframes, and the estimated velocity of the robot/sensor is also used to compensate for motion distortion point clouds extracted from radar data. Lastly, CFEAR-3~\cite{adolfsson2023cfear3} proposed a combination of weighting, keyframe history, filtering, and a cost function that overcomes the previous challenges with sparsity, bias, and overly conservative filtering, making the method the current state-of-the-art in radar odometry.

Alhashimi et al.~\cite{alhashimi2021bfar} presented a method built on the CFEAR framework, by using a feature extraction algorithm called Bounded False Alarm Rate (BFAR) to add a constant offset to the usual Constant False Alarm Rate (CFAR) threshold. The resulting radar point clouds are registered to a sliding window of keyframes using an ICP-like optimizer.




\begin{figure*}[!t]
    \vspace{.1em} 
    \centering
    \includegraphics[width=0.99\textwidth]{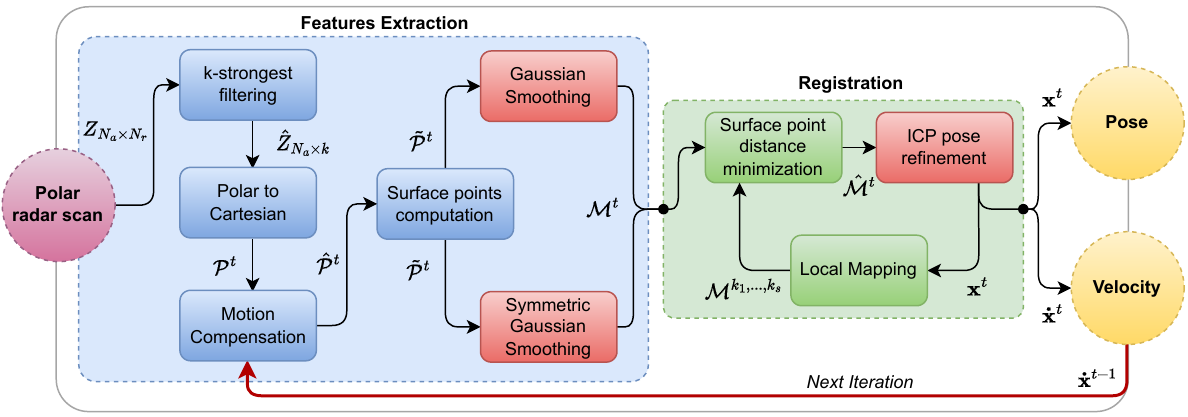}
    \caption{Our pipeline enhances radar odometry by incorporating a series of steps: filtering polar radar scan, motion compensation, calculating oriented surface points, smoothing, registering multiple radar scans at once, and refining the pose through ICP. The process leverages smoothing techniques to supplement additional information and aligns successive scans for further refinement after the bulk registration.}
    \label{fig:pipeline}
\end{figure*}

Over the years, a few radar-based SLAM systems have been proposed. The work in~\cite{holder2019real} proposed an ICP-based graph-based SLAM system consisting of preprocessing, odometry estimation via data fusion, scan matching, loop closure, graph construction, and periodic optimization.



Recently, the same authors of CFEAR presented also TBV (Trust But Verify) Radar SLAM~\cite{adolfsson2023tbv}, a method for radar SLAM that verifies loop closure candidates and uses CFEAR-3 as odometry estimation method.
The authors built on the place recognition method Scan Context by Kim et al.~\cite{kim2018scan, kim2020mulran}, which detects loops and relative orientation by matching descriptors associated with multiple point clouds.

\section{Radar odometry estimation}\label{section:pipeline}
Our incremental odometry estimation method, represented in Fig.~\ref{fig:pipeline}, builds upon CFEAR-3~\cite{adolfsson2023cfear3}, operating on data gathered by a spinning radar, represented in polar form. Sensor data is first processed to keep only the k-strongest readings per azimuth, to filter out noise points. The filtered point cloud is then compensated for motion distortion using the estimated velocity from the previous iteration. Sparse-oriented surface points are then computed using a grid-based approach that optionally smooths the point distribution around each grid cell. The pose and velocity of the radar sensor are computed by registering the current scan to a history of previous, non-consecutive, scans (i.e., keyframes, forming a local map) and the pose is optionally refined through an ICP-based scan matching solution.

\subsection{Radar pre-filtering} 
Given a polar radar scan of size $N_{a} \times N_{r}$, with $N_{a}$ azimuth and $N_{r}$ range bins, radar pre-filtering iterates over all azimuth bins separately. For each azimuth, we choose the k-range bins that have the highest power returns, meaning they have intensities in the polar image that surpass the expected noise level, to reduce common data artifacts.

After filtering, each selected range and azimuth pair $(d,a)$, with $d\in [1,\dots,N_{r}]$ and $a\in [1,\dots,N_{a}]$, is transformed to a point in the Cartesian space, for subsequent algorithms:
\begin{equation}
    \textbf{p} = \begin{bmatrix} p_{x} \\ p_{y} \end{bmatrix} = \begin{bmatrix} d\, \gamma\, cos(\theta) \\ d\, \gamma\, sin(\theta) \end{bmatrix},
\end{equation}

where $\theta = 2\pi a / N_{a}$ and $\gamma$ is the sensor range resolution. The output point cloud is the set of all Cartesian points.

\subsection{Motion compensation}
Considering the most recent linear and angular estimated velocities $\dot{\mathbf{x}}_{t-1} = \begin{bmatrix} v_{x}&v_{y}&\omega \end{bmatrix}$, the filtered point cloud is compensated for motion by projecting each point into the time $t$ of the center of the radar sweep, such that $t_{i} = t + \delta t$, where $\delta t = \begin{bmatrix} -\Delta T / 2, \Delta T / 2 \end{bmatrix}$ and $\Delta T$ is the duration of a full sweep. The time offset $\delta t$ of each point is then computed from the index $a_{\Delta t}$ of the corresponding azimuth bin and number of bins per sweep: $\delta t = (a_{\Delta t} - N_{a} / 2)\, \Delta T /2$.
Lastly, each point $p_{\delta t}$ can be corrected accordingly to the inverse of the distortion terms, compensating for its motion.

\subsection{Oriented surface points computation}
Instead of considering the filtered and motion-compensated point cloud as presented, we conceptualize it as an assembly of oriented surface points $\mathcal{M}_{t}=\{m_{i}\}$, to sparsely model geometries and data distribution within the scene. Each oriented surface point consists of mean and normal values, i.e., $m_{i}=(\mu_{i}, \mathbf{n}_{i})$. The corrected point cloud is downsampled using a regular grid with cell size $r$.

For each cell in the grid that has at least one point, all $l$ points within a radius $r$ from the cell centroid, represented as matrix $\mathbf{P}_{2\times l}$, are used to compute the weighted sample mean $\mu_{i}$ and weighted sample covariance:

\begin{equation}
    \mathbf{\Sigma_{i}} = (\mathbf{P} - \begin{bmatrix} \mu_{i} \end{bmatrix}_{\times l})\mathbf{W}(\mathbf{P} - \begin{bmatrix} \mu_{i} \end{bmatrix}_{\times l})^{T},
\end{equation}
where $\mathbf{W}$ is a normalized diagonal matrix whose elements are proportional to the reflected intensity, as in~\cite{kung2021normal}:
\begin{equation}
    \mathbf{W}_{j,j}=z_{j} - z_{min},
\end{equation}


The search radius $r$ determines the final representation’s target density of the environment, which is the set of oriented surface points. Considering that this parameter shares a value with the grid cell size, it allows a single point to contribute to the calculation of multiple surface points. 

To improve the computation of these points, we incorporated two additional and mutually exclusive enhancements, which rely on smoothing techniques, described as follows.

\textbf{Gaussian kernel Smoothing.} Rather than directly utilizing the mean and covariance of each oriented surface point, a smoothing kernel can be applied to the grid outlined in the preceding sections. This allows both values to be considered representative of each grid cell. Consequently, the distribution maintained by a single cell not only is associated with the distribution of points within a radius $r$ from its center, but it also represents the weighted distributions of points belonging to the surrounding cells.

Given a typical $3\times 3$ Gaussian kernel, represented as
\begin{equation*}
    \text{GK} = \frac{1}{16} \begin{bmatrix} 1&2&1 \\ 2&4&2 \\ 1&2&1 \end{bmatrix},
\end{equation*}
where each element can be seen as a distribution importance weight $w_{i}$, left to right, top to bottom, the new mean and covariance for each surface point become:
\begin{equation}
    \begin{aligned}
        \hat{\mu}_{i} &= \sum_{i \in \{1,\dots,9\}} w_{i} \mu_{i}, \\
        \hat{\mathbf{\Sigma}}_{i} &= \sum_{i \in \{1,\dots,9\}} w_{i} (\mathbf{\Sigma_{i}} + \mu_{i} \mu_{i}^{T}) - \hat{\mu}_{i} \hat{\mu}_{i}^{T}.
    \end{aligned}
\end{equation}

It is essential to notice that each weight $w_{i}$ is also multiplied by the number of points $l$ that were used to compute the original mean and covariance of each cell before the sum of weights is normalized to $1$. This smoothing method can be applied without incurring any additional storage or computational expenses.

\textbf{Symmetric Gaussian kernel Smoothing.} A potential drawback of Gaussian kernel smoothing is that in non-uniform environments, such as urban settings, the distribution of points may be inconsistent across neighboring cells. Consequently, the smoothing process could alter the mean and covariance, skewing these values towards cells with higher density. This could lead to issues during the registration process, such as incorrect or misleading correspondences.

The solution we propose, as an alternative to a simple Gaussian kernel, is to use \textit{symmetric smoothing}. Let $W_{k}$ be the weights matrix of size $k \times k$, already normalized considering the number of points used to compute the various mean and covariance values $\mu_{i}$ and $\mathbf{\Sigma_{i}}$. Before we proceed with the smoothing process, we ensure that the weights are \textit{symmetric} with respect to the central element. If there exists at least one pair where one element is positive and the other is zero, then we abstain from performing smoothing. 

Specifically, smoothing is only carried out if all weights in the kernel satisfy either of the given conditions:
\begin{equation}
    \begin{aligned}
        &(w_{i,j}>0) \land (w_{k-i, k-j}>0), \\
        &w_{i,j} = w_{k-i, k-j}=0.
    \end{aligned}
\end{equation}

As a last step, eigen-decomposition on the sample covariance $\mathbf{\Sigma}_{i}$ allows to estimate the surface normal $\mathbf{n}_{i}$, using the eigenvector that corresponds to the smallest eigenvalue. Where mean and normal have been successfully computed, the oriented surface point is then obtained as $m_{i}=(\mu_{i}, \mathbf{n}_{i})$.

\subsection{Local mapping}
Keyframes, which are composed of oriented surface points known as reference scans and the corresponding robot poses, are stored in a sliding window of size $s$ at regular intervals. This only occurs when the pose, estimated through registration, surpasses a certain threshold distance in either rotation or translation from the previous keyframe. This method prevents unnecessary computation and memory usage when the robot’s movement is slow or stationary, such as when waiting at an intersection, forming efficient local submaps.


\subsection{Registration}
The pose of the robot is estimated by finding optimization parameters $\textbf{x}_t = \begin{bmatrix} x&y&\theta \end{bmatrix}^T$ that best align the latest scan $\mathcal{M}_t$ with a window of the most recent keyframes $\mathcal{M}_i, i \in \mathcal{K}=\{1,\dots,K\}$, with $K$ being a fixed window size, according to the function to minimize:
\begin{equation}
    \textit{f}(\mathcal{M}_{\mathcal{K}}, \mathcal{M}_{t}, \textbf{x}_{t}) = \sum_{k \in \mathcal{K}} \sum_{(i,j)\in \mathcal{C}} w_{i,j}\mathcal{L}_{\delta}(g(m_{k_{j}}, m_{t_{i}}, \textbf{x}_{t})).
\end{equation}

The term $\mathcal{C}$ is the set of correspondences between nearby oriented surface points associated with two keyframes, found by radius search and surface normal similarity.

\begin{figure*}[!t]
    \centering
    \vspace{.3em} 
    \includegraphics[width=0.32\textwidth]{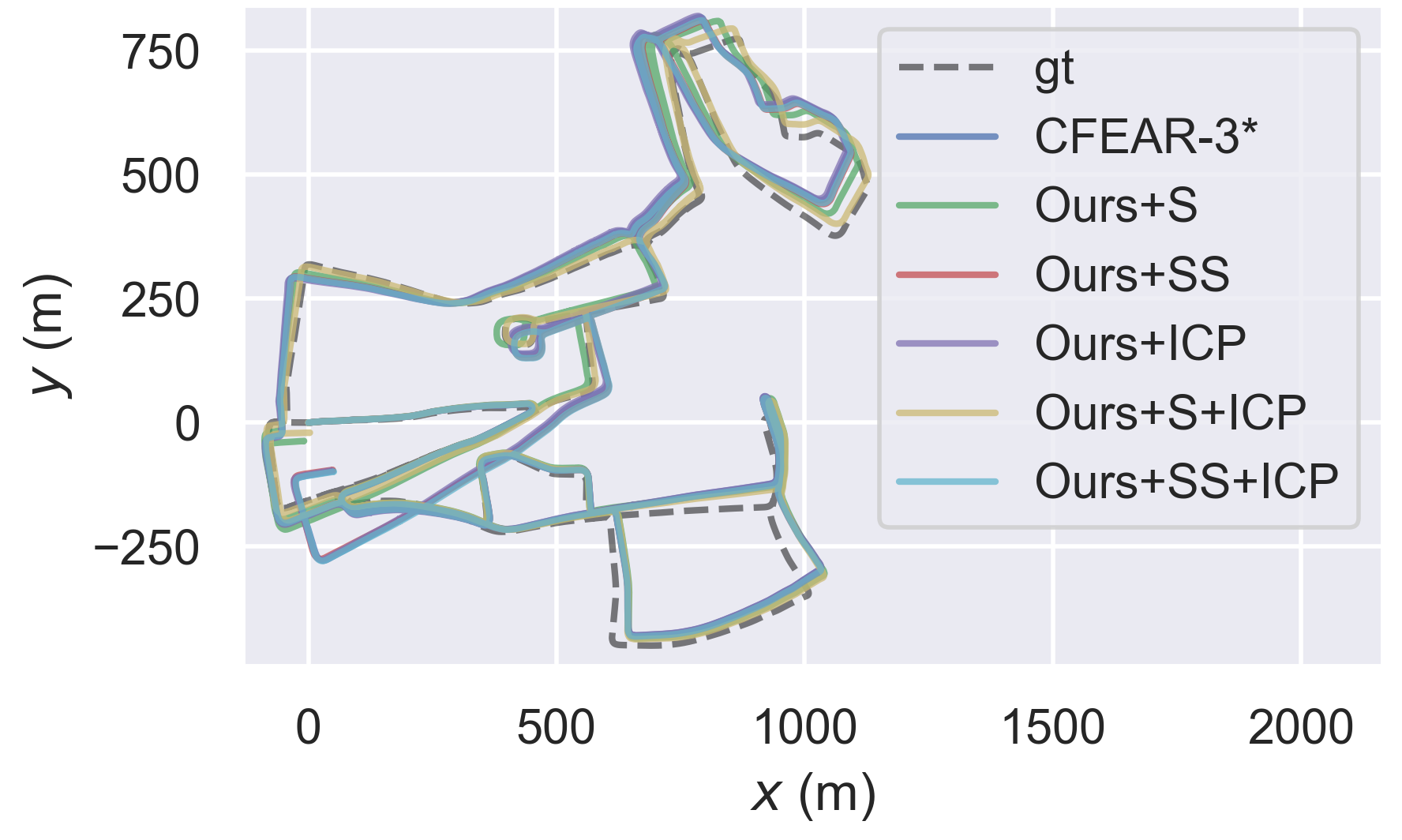}
    \includegraphics[width=0.3\textwidth]{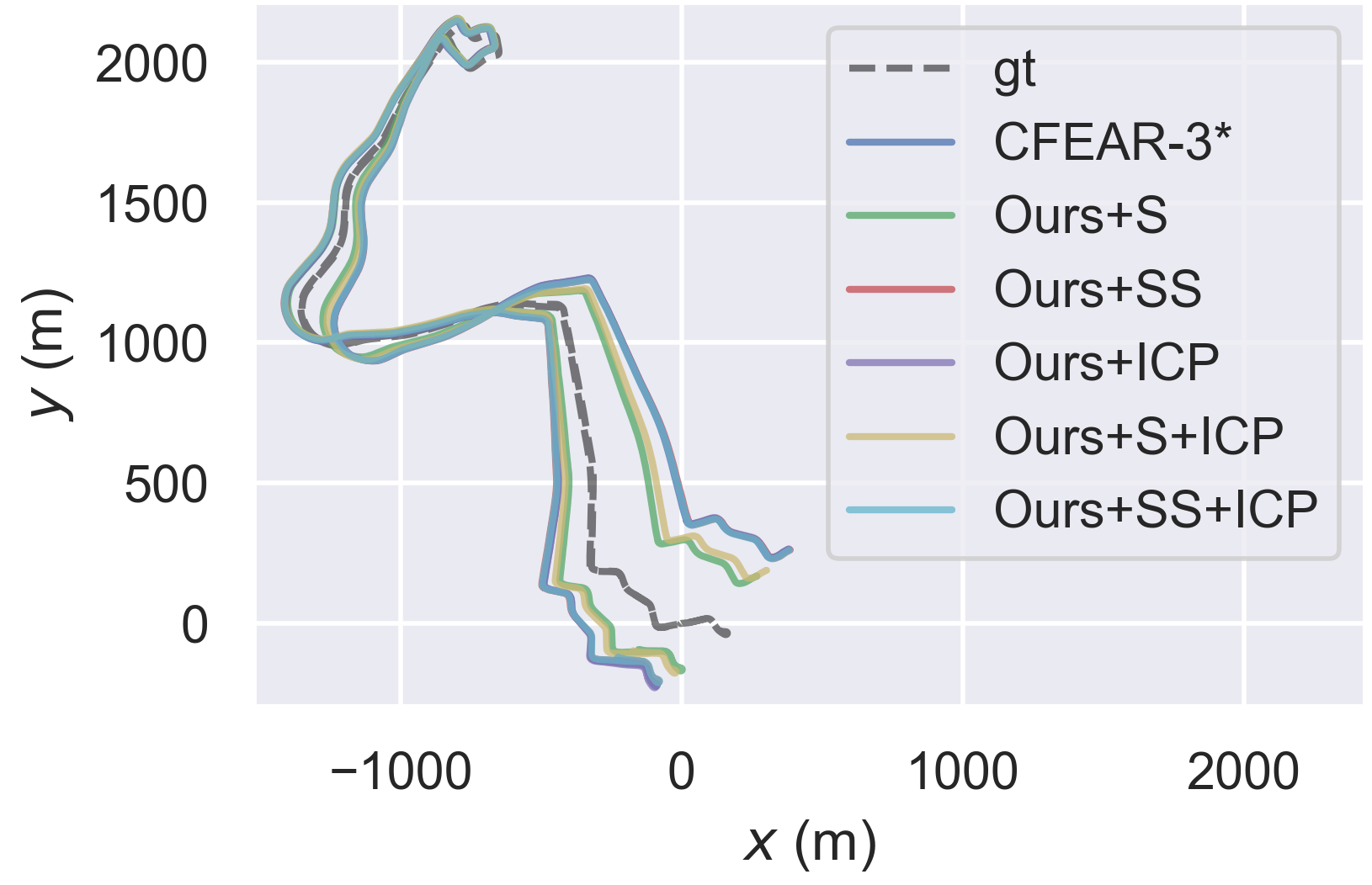}
    \includegraphics[width=0.34\textwidth]{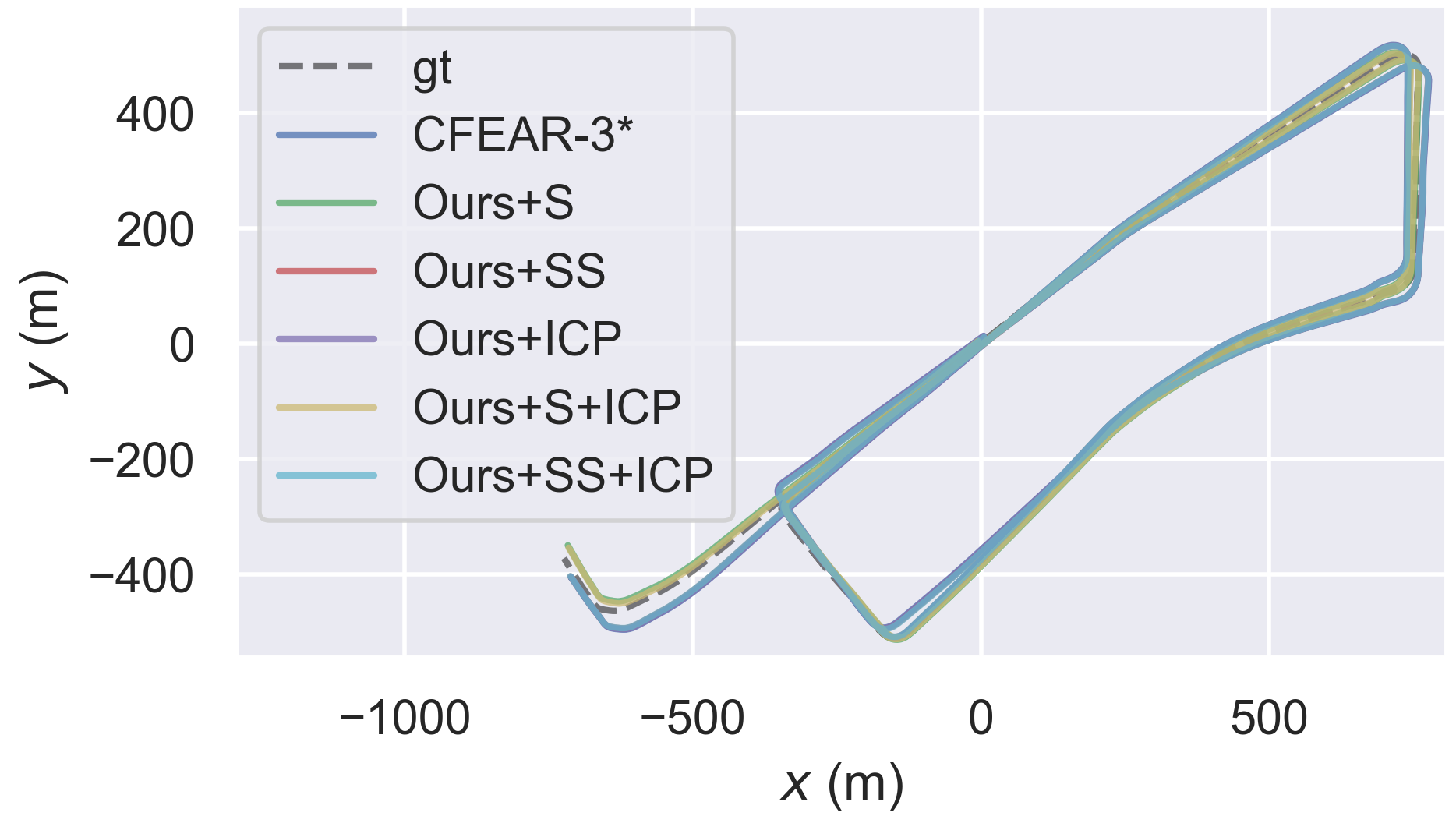}
    \caption{Odometry trajectory comparison among our pipeline settings with respect to CFEAR-3* and the ground truth. Starting from left to right, the first image is Oxford 18-15-20; the second image is Boreas 01-26-11: and the last image is DCC01.}
    \label{figure:trajectories}
\end{figure*}

As the cost function, we consider the point-to-point (P2P) approach~\cite{arun1987least}, which proves to be the most reliable:
\begin{equation}
    g_{P2P}(m_{k_{j}}, m_{t_{i}}, \textbf{x}_{t}) = \lVert \mathbf{e} \rVert^{2},
    \label{equation:registration}
\end{equation}
with $\mathbf{e}=\mu_{k_{j}} - (\mathbf{R}_{\theta}\mu_{t_{i}} + \mathbf{t}_{x,y})$ being the error term w.r.t. the translation and rotation (matrix form) components of $\textbf{x}_{t}$.

The symbol $\mathcal{L}_{\delta}$ denotes the \textit{Huber Loss}, which is designed to make the cost function less influenced by outliers. 

Finally, all residuals are adjusted based on the similarity of surface points, taking into account factors such as point planarity, the number of detections used to calculate the points, and the direction of normals:
\begin{equation}
    \begin{aligned}
        w_{i,j} &= w_{i,j}^{plan} + w_{i,j}^{dim} + w_{i,j}^{dir} \\
        &= f_{sim}(p_{i},p_{j}) + f_{sim}(n_{i},n_{j}) + \text{max}(\mathbf{n}_{i}\cdot \mathbf{n}_{j}, 0),
    \end{aligned}
\end{equation}
where $f_{sim}(a,b)=2\,\text{min}(a,b)/(a+b)$ represents the similarity between two values, and $p=\text{log}(1+|\lambda_{max} / \lambda_{min}|)$ is the planarity of a surface point, with $\lambda_{max}$ and $\lambda_{min}$ being the eigenvalues of the corresponding covariance matrix.



Intuitively, this procedure enforces spatial and temporal consistency between radar scans. Nevertheless, the representation and registration based on oriented surface points may not use some point-based information, which can be instead exploited to make odometry estimates more accurate.

\textbf{ICP-based pose refinement.} In a similar manner to our approach for calculating oriented surface points, where we implemented enhancements (including Gaussian kernel and symmetric Gaussian kernel smoothing) over CFEAR-3, we have also expanded the registration process with an optional pose refinement feature, following the previous step. 

First, we compute the mean values of each set of oriented surface points, obtaining their corresponding positions in the Cartesian space. Given all the derived points, we form a 2D point cloud representing the various power returns of each radar scan. The point cloud corresponding to the current radar image is then aligned to the point cloud derived from the previous radar scan, not to be mistaken with the previous keyframe, which may correspond to a few images before the current one (depending on the traveled distance). To perform this alignment, we use the robot's pose estimated at the previous step of the pipeline, meaning the scan-to-multi keyframe registration, as initial guess, due to its overall accuracy. After this ICP-based alignment, we compute the residuals of the various point-to-point correspondences. If their sum is below a stringent threshold (i.e., an Euclidean fitness valued below 1.0), the pose is updated with the new estimate resulting from the alignment. 

This method offers the benefit of further refining consecutive robot steps by ensuring both temporal and spatial consistencies in small windows (due to the consecutive nature of the alignment). This way, we are able to consider the oriented surface points as a whole point cloud, allowing for a more precise pose estimation.

\section{Evaluation}
\subsection{Setup}
We conducted an extensive experimental analysis on the major three datasets within the radar literature:
\begin{itemize}
    \item Oxford Radar RobotCar~\cite{barnes2020oxford} (Navtech CTS350-X);
    \item MulRan~\cite{kim2020mulran} (Navtech CIR204-H);
    \item Boreas~\cite{burnett_ijrr23} (Navtech CIR304-H).
\end{itemize}
Both Oxford and MulRan had been fully tested on all sequences, while not all trajectories of Boreas had a ground truth to use for comparison. As far as we know, this is the first paper in which Boreas has been used as dataset for evaluating multiple algorithms for radar odometry.

Each radar sensor model has been set according to the Navtech datasheet with respect to its range resolution, while hyper-parameters of the pipeline have been chosen according to the CFEAR-3 values reported by Adolfsson et al.~\cite{adolfsson2023cfear3}. It is also important to notice that the Boreas sequences were tested using two different range and resolution values, as the authors state in~\cite{burnett_ijrr23}, due to a sudden driver update. 

We compare our baseline, which is our own C++ re-implementation of CFEAR (referred to as CFEAR-1*, CFEAR-2*, and CFEAR-3*, respectively, depending on the configuration of parameters), with all possible combinations of the improvements detailed in Section~\ref{section:pipeline}, and other methods in the literature. These improvements include Gaussian kernel smoothing (S), Symmetric Gaussian kernel smoothing (SS), and ICP-based pose refinement (ICP). We also consider combinations of these improvements, such as Gaussian kernel smoothing with ICP (S + ICP) and Symmetric Gaussian kernel smoothing with ICP (SS + ICP). Fig.~\ref{figure:trajectories} shows a visual comparison of the proposed pipeline with the various improvements, over multiple trajectories.

Moreover, we distinguish between paper and GitHub versions of CFEAR-1* and CFEAR-2*, as the used parameters slightly differ. This distinction is not done for CFEAR-3*, as the set of parameters is unique (refer to~\cite{adolfsson2023cfear3}).

The comparison is achieved through multiple metrics, i.e., translation error percentage, rotation error expressed in degrees over 100 meters, relative pose error (RPE), in centimeters, and absolute trajectory error (ATE), in meters. The metrics are computed using a well-known and existing benchmarking toolbox~\footnote{https://github.com/dan11003/radar\_kitti\_benchmark/tree/master}. We do not include the translation error percentage and rotation error for the Boreas dataset, as the obtained values are considerably high and not as meaningful as the other metrics (i.e., RPE and ATE).

\subsection{Qualitative and Quantitative Results}

\subsubsection{Oxford Radar RobotCar}

\begin{table*}[ht]
\begin{center}
\vspace{.3em} 
\def\arraystretch{1.1}
\begin{adjustbox}{width=1\textwidth}
\begin{tabular}{l|c c c c c c c c |c c}
    \% translation error / deg/100 [m] & \textit{Sequence} & & & & & & & & & \\
    \hline
    Method & 10-12-32 & 16-13-09 & 17-13-26 & 18-14-14 & 18-15-20 & 10-11-46 & 16-11-53 & 18-14-46 & Mean & Mean$^{\dag}$\\
    \hline
    Cen2018~\cite{cen2018precise} & - & - & - & - & - & - & - & - & 3.72/0.95 & - \\ 
    MC-RANSAC~\cite{burnett2021we} & - & - & - & - & - & - & - & - & 3.31/1.09 & - \\
    CC-MEANS~\cite{aldera2022goes} & - & - & - & - & - & - & - & - & 2.53/0.82 & - \\
    RadarSLAM (odom)~\cite{hong2022radarslam} & 2.32/0.7 & 2.62/0.7 & 2.27/0.6 & 2.29/0.7 & 2.25/0.7 & 2.16/0.6 & 2.49/0.7 & 2.12/0.6 & 2.32/0.7 & - \\
    Kung~\cite{kung2021normal} & - & - & - & - & 2.20/0.77 & - & - & - & 1.96/0.6 & - \\
    CFEAR-1~\cite{adolfsson2021cfear1} & 1.59/0.57 & 1.84/0.64 & 1.84/0.64 & 1.83/0.60 & 1.71/0.59 & 1.74/0.57 & 2.11/0.63 & 1.69/0.54 & 1.79/0.60 & - \\
    CFEAR-2~\cite{adolfsson2021cfear2} & 1.35/0.49 & 1.50/0.51 & 1.52/0.54 & 1.52/0.52 & 1.41/0.50 & 1.33/0.48 & 1.61/0.53 & 1.48/0.50 & 1.46/0.51 & - \\
    CFEAR-3~\cite{adolfsson2023cfear3} & 1.23/0.36 & 1.25/0.39 & 1.25/0.40 & 1.34/0.41 & 1.26/0.41 & 1.26/0.39 & 1.42/0.39 & 1.42/0.44 & 1.31/0.40 & - \\
    \hline 
    CFEAR-1*$_{paper}$ & 1.83/0.62 & 2.28/0.75 & 2.24/0.80 & 2.14/0.72 & 2.19/0.76 & 1.99/0.73 & 2.62/0.75 & 2.11/0.73 & 2.18/0.73 & 2.20/0.74\\
    CFEAR-1*$_{github}$ & 1.81/0.62 & 1.82/0.64 & 2.01/0.71 & 2.14/0.70 & 1.93/0.70 & 1.90/0.67 & 1.97/0.70 & 2.15/0.74 & 1.94/0.69 & 1.93/0.67\\
    CFEAR-2*$_{paper}$ & 1.76/0.57 & 1.88/0.63 & 1.89/0.68 & 1.92/0.64 & 1.64/0.58 & 1.67/0.60 & 1.78/0.60 & 1.80/0.62 & 1.79/0.62 & 1.77/0.61\\
    CFEAR-2*$_{github}$ & 1.63/0.57 & 1.69/0.60 & 1.78/0.64 & 1.89/0.65 & 1.77/0.64 & 1.60/0.57 & 1.75/0.62 & 1.88/0.66 & 1.75/0.62 & 1.71/0.60\\
    \hline
    CFEAR-3* & \textbf{1.29}/0.36 & 1.35/\textbf{0.39} & \textbf{1.30}/\textbf{0.39} & 1.43/0.42 & 1.40/0.43 & \textbf{1.33}/0.40 & 1.49/0.41 & \textbf{1.46}/\textbf{0.42} & 1.38/\textbf{0.40} & 1.37/0.41 \\
    CFEAR-3*$_{S}$ & 2.21/0.47 & 2.45/0.49 & 2.05/0.56 & 1.72/\textbf{0.41} & 1.70/\textbf{0.41} & 1.52/0.40 & 2.18/0.43 & 1.75/0.43 & 1.95/0.45 & 1.95/0.44 \\
    CFEAR-3*$_{SS}$ & 1.30/0.37 & 1.33/0.40 & 1.31/0.40 & 1.43/0.42 & 1.36/0.42 & 1.34/0.39 & 1.48/0.41 & 1.47/0.43 & 1.39/0.41 & 1.36/\textbf{0.40} \\
    CFEAR-3*$_{ICP}$ & 1.29/\textbf{0.36} & 1.35/0.40 & 1.30/0.39 & 1.43/0.42 & 1.38/0.43 & 1.33/0.40 & 1.50/0.41 & 1.46/0.43 & \textbf{1.38}/0.41 & 1.37/0.40 \\
    CFEAR-3*$_{S + ICP}$ & 1.73/0.42 & 1.79/0.44 & 1.66/0.45 & 1.72/0.41 & 1.64/0.42 & 1.54/0.41 & 2.11/0.43 & 1.76/0.43 &  1.73/0.43 & 1.84/0.43 \\
    CFEAR-3*$_{SS + ICP}$ & 1.30/0.36 & \textbf{1.33}/0.40 & 1.31/0.40 & \textbf{1.43}/0.42 & \textbf{1.36}/0.42 & 1.33/\textbf{0.39} & \textbf{1.48}/\textbf{0.41} & 1.46/0.43 & 1.39/0.40 & \textbf{1.36}/0.40\\
    \hline
\end{tabular}
\end{adjustbox}
\end{center}
\caption{Comparison in \% translation error and deg/100m on the Oxford Radar RobotCar dataset~\cite{barnes2020oxford}. \textit{(*)} is our re-implementation; \textit{(S)} is smoothing; \textit{(SS)} is symmetric smoothing; \textit{(ICP)} is scan-matching; \textit{(paper)} refers to the parameters as described in~\cite{adolfsson2023cfear3}; \textit{(github)} refers to the parameters as described in the Github repository of CFEAR-3; and (\dag) is the mean over all sequences.}
\label{table:oxford_tr_deg}
\end{table*}

\begin{table*}[ht]
\begin{center}
\def\arraystretch{1.1}
\begin{adjustbox}{width=1\textwidth}
\begin{tabular}{l|c c c c c c c c |c c}
    RPE [cm] / ATE[m] & \textit{Sequence} & & & & & & & & & \\
    \hline
    Method & 10-12-32 & 16-13-09 & 17-13-26 & 18-14-14 & 18-15-20 & 10-11-46 & 16-11-53 & 18-14-46 & Mean & Mean$^{\dag}$\\
    \hline
    CFEAR-1*$_{paper}$ & 5.57/49.38 & 6.09/45.58 & 5.44/104.26 & 5.82/71.41 & 5.51/63.89 & 5.83/82.35 & 6.57/53.11 & 5.75/60.38 & 5.82/66.42 &6.08/65.86\\
    CFEAR-1*$_{github}$ & 4.78/56.86 & 5.28/60.42 & 4.40/63.00 & 5.10/62.44 & 4.74/72.77 & 5.08/59.92 & 5.59/69.83 & 5.06/79.57 & 4.88/65.60 &5.25/58.72\\
    CFEAR-2*$_{paper}$ & 5.34/32.37 & 5.86/55.15 & 5.06/70.21 & 5.63/57.85 & 5.28/52.25 & 5.55/44.33 & 6.04/40.58 & 5.61/56.71 & 5.42/51.05 &5.77/46.96\\
    CFEAR-2*$_{github}$ & \textbf{4.66}/42.00 & \textbf{5.17}/52.16 & \textbf{4.26}/56.03 & \textbf{5.05}/62.69 & \textbf{4.67}/57.47 & \textbf{5.01}/36.96 & \textbf{5.47}/50.81 & \textbf{4.94}/73.35 & \textbf{4.78}/53.81 &\textbf{5.15}/47.31\\
    \hline
    CFEAR-3* & 5.41/\textbf{5.29} & 5.82/\textbf{20.43} & 5.10/11.92 & 5.55/15.26 & 5.14/25.55 & 5.64/13.71 & 5.80/15.44 & 5.46/23.59 & 5.37/16.40 & 5.71/17.76\\
    CFEAR-3*$_{S}$ & 10.24/50.04 & 10.54/66.35 & 9.75/89.55 & 8.60/12.07 & 8.26/9.57 & 9.13/20.84 & 9.31/19.70 & 8.20/11.30 & 9.26/34.90 &9.40/27.56\\
    CFEAR-3*$_{SS}$ & 5.54/6.03 & 5.94/23.43 & 5.21/\textbf{9.95} & 5.66/17.25 & 5.23/25.10 & 5.78/\textbf{9.58} & 5.94/15.40 & 5.53/25.32 & 5.60/16.51 &5.83/\textbf{17.61}\\
    CFEAR-3*$_{ICP}$ & 5.96/5.36 & 6.04/20.49 & 5.36/11.67 & 5.89/15.90 & 5.78/24.62 & 6.28/12.21 & 6.27/\textbf{14.11} & 5.92/23.53 & 5.94/\textbf{16.36} &6.13/17.65\\
    CFEAR-3*$_{S + ICP}$ & 10.08/17.94 & 9.91/28.16 & 9.66/23.88 & 8.90/\textbf{10.36} & 8.70/\textbf{9.06} & 9.78/23.48 & 9.77/19.56 & 8.55/\textbf{10.16} & 9.54/17.95 &9.62/21.11\\
    CFEAR-3*$_{SS + ICP}$ & 6.11/5.98 & 6.14/23.73 & 5.46/10.03 & 6.01/18.30 & 5.88/26.12 & 6.41/9.73 & 6.37/15.37 & 5.95/24.83 & 6.04/16.64 &6.23/17.71\\
    \hline
\end{tabular}
\end{adjustbox}
\end{center}
\caption{RPE [cm] and ATE [m] metrics, respectively, on the Oxford Radar RobotCar dataset~\cite{barnes2020oxford}. \textit{(*)} is our re-implementation; \textit{(S)} is smoothing; \textit{(SS)} is symmetric smoothing; \textit{(ICP)} is scan-matching; \textit{(paper)} refers to the parameters as described in~\cite{adolfsson2023cfear3}; \textit{(github)} refers to the parameters as described in the Github repository of CFEAR-3; and (\dag) is the mean over all sequences.}
\label{table:oxford_rpe_ate}
\end{table*}

In this dataset, despite the lack of explicit weather conditions, it was possible to infer them from the camera recordings. However, most of the recordings were made under overcast weather conditions, which did not allow for clear qualitative conclusions. 

From a quantitative standpoint, Table~\ref{table:oxford_tr_deg} shows that the proposed improvements to the pipeline do not lead to better results when considering the sequences that are always used for odometry evaluation in the literature. Nevertheless, when studying the translation and rotation error over all sequences, it becomes clear that the developed methods achieve increased accuracy, having ICP-based solutions performing better than all other alternatives, in translation. 

We can also assert that the proposed improvements outperform CFEAR-3* on average across the recorded scenes, with significant improvements in ATE, expressed in Table~\ref{table:oxford_rpe_ate}), observed in specific local scenes such as 18-14-46 or 10-11-46, among others. Of particular interest is the RPE metric, which achieves minimum values always when derived from the CFEAR-2* method using the parameters as described in the corresponding GitHub repository.

Lastly, it can be noticed that the Gaussian kernel smoothing leads to worse results. This is most likely related to the type of environment represented in the dataset: unbalanced scenes strongly bias the smoothing kernel, decreasing the overall odometry estimation accuracy. 

\subsubsection{MulRan }

\begin{table*}[ht]
\begin{center}
\def\arraystretch{1.1}
\begin{adjustbox}{width=1\textwidth}
\begin{tabular}{l|c c c | c c c  | c c c  | c c c |c c}
    & \multicolumn{2}{l}{\textit{Sequence}} & \multicolumn{1}{c}{} & & & \multicolumn{1}{c}{} & & & \multicolumn{1}{c}{} & & & \multicolumn{1}{c|}{} & &
    \\
    \hline
    \multirow{2}{*}{Method} & \multicolumn{3}{c|}{DCC} & \multicolumn{3}{c|}{KAIST} & \multicolumn{3}{c|}{Riverside} & \multicolumn{3}{c|}{Sejong} & \multirow{2}{*}{Mean$^{\dag}$} & \multirow{2}{*}{Mean}\\
     & 01 & 02 & \multicolumn{1}{c|}{03} & 01 & 02 & 03 & 01 & 02 & 03 & 01 & 02 & 03 & & \\
    \hline
    RadarSLAM (odom)~\cite{hong2022radarslam} & 2.70/0.5 & 1.90/0.4 & 1.64/0.4 & 2.13/0.7 & 2.07/0.6 & 1.99/0.5 & 2.04/0.5 & 1.51/0.5 & 1.71/0.5 & - & - & - & 1.97/0.5 & -\\
    CFEAR-1~\cite{adolfsson2021cfear1} & 2.73/0.73 & 1.82/0.60 & 1.77/0.62 & 2.62/0.97 & 2.45/0.90 & 2.85/1.08 & 2.55/0.90 & 2.71/0.82 & 3.56/0.82 & - & - & - & 2.56/0.83 & -\\
    CFEAR-2~\cite{adolfsson2021cfear2} & 2.44/0.63 & 1.65/0.54 & 1.41/0.50 & 2.12/0.81 & 1.93/0.74 & 2.08/0.87 & 2.30/0.80 & 2.07/0.66 & 2.60/0.59 & - & - & - & 2.07/0.68 & -\\
    CFEAR-3~\cite{adolfsson2023cfear3} & 2.28/0.54 & 1.49/0.46 & 1.47/0.48 & 1.59/0.66 & 1.62/0.66 & 1.73/0.78 & 1.59/0.63 & 1.39/0.51 & 1.41/0.40 & - & - & - & 1.62/0.57 & -\\
    \hline
    CFEAR-1*$_{paper}$ & 2.89/0.68 & 2.18/0.56 & 1.78/0.50 & 3.03/0.95 & 2.02/0.66 & 2.50/0.79 & 2.52/0.72 & 2.56/0.73 & 4.00/0.79 & 2.98/0.63 & 3.55/0.85 & 3.63/0.83 & 2.80/0.72 & 2.61/0.71\\
    CFEAR-1*$_{github}$ & 2.55/0.55 & 1.65/0.41 & 1.51/0.43 & 2.16/0.68 & 1.71/0.52 & 1.65/0.52 & 2.43/0.69 & 2.21/0.65 & 2.51/0.52 & 2.38/0.62 & 2.98/0.70 & 3.12/0.77 & 2.24/0.59 & 2.04/0.55\\ 
    CFEAR-2*$_{paper}$ & 2.41/0.50 & 1.58/0.39 & \textbf{1.16}/0.31 & 1.75/0.53 & 1.64/0.50 & 1.70/0.52 & 2.18/0.60 & 1.96/0.54 & 2.64/0.52 & 3.11/0.53 & 2.79/0.65 & 2.73/0.68 & 2.14/0.52 & 1.89/0.49\\
    CFEAR-2*$_{github}$ & 2.41/0.49 & 1.54/0.38 & 1.21/0.31 & 1.63/0.51 & 1.54/0.46 & 1.62/0.50 & 2.12/0.58 & 1.90/0.53 & 2.53/0.49 & 3.80/0.53 & 2.82/0.66 & 2.77/0.69 & 2.16/0.51 & 1.83/0.47\\
    \hline
    CFEAR-3* & 2.12/\textbf{0.29} & 1.29/0.22 & 1.29/\textbf{0.26} & \textbf{1.13}/0.32 & \textbf{1.23}/\textbf{0.37} & 1.18/0.34 & \textbf{1.21}/\textbf{0.35} & 1.18/0.31 & 1.36/0.23 & 1.10/0.24 & \textbf{1.58}/\textbf{0.32} & 1.65/0.42 & 1.36/\textbf{0.30} & 1.33/0.30 \\
    CFEAR-3*$_{S}$ & 2.37/0.33 & 1.64/0.26 & 1.77/0.31 & 1.34/0.35 & 1.39/0.40 & 1.31/0.36 & 1.57/0.38 & 1.78/0.37 & 1.64/\textbf{0.21} & 1.48/0.25 & 1.91/0.34 & 1.93/0.43 & 1.68/0.33 & 1.64/0.33\\
    CFEAR-3*$_{SS}$ & 2.13/0.29 & 1.31/0.22 & 1.32/0.27 & 1.14/0.32 & 1.25/0.37 & \textbf{1.16}/\textbf{0.33} & 1.24/0.35 & 1.15/0.31 & 1.37/0.23 & 1.10/0.24 & 1.61/0.33 & \textbf{1.64}/\textbf{0.42} & 1.37/0.31 & 1.34/0.30\\
    CFEAR-3*$_{ICP}$ & \textbf{2.12}/0.29 & \textbf{1.29}/0.22 & 1.29/0.26 & 1.13/\textbf{0.32} & 1.23/0.37 & 1.18/0.34 & 1.21/0.35 & 1.18/0.31 & \textbf{1.36}/0.23 & 1.11/0.24 & 1.59/0.32 & 1.65/0.42 & \textbf{1.36}/0.31 & \textbf{1.33}/\textbf{0.30} \\
    CFEAR-3*$_{S + ICP}$ & 2.37/0.33 & 1.64/0.26 & 1.77/0.31 & 1.34/0.32 & 1.39/0.40 & 1.31/0.36 & 1.56/0.38 & 1.78/0.37 & 1.65/0.21 & 1.48/0.24 & 1.90/0.34 & 1.93/0.43 & 1.68/0.33 & 1.65/0.33\\
    CFEAR-3*$_{SS + ICP}$ & 2.13/0.29 & 1.31/\textbf{0.22} & 1.31/0.27 & 1.14/0.32 & 1.25/0.37 & 1.16/0.33 & 1.24/0.35 & \textbf{1.15}/\textbf{0.31} & 1.37/0.23 & \textbf{1.10}/\textbf{0.24} & 1.61/0.32 & 1.64/0.42 & 1.37/0.31 & 1.34/0.30\\
    \hline
\end{tabular}
\end{adjustbox}
\end{center}
\caption{Comparison in \% translation error and deg/100m on the MulRan dataset~\cite{kim2020mulran}. \textit{(*)} is our re-implementation of CFEAR; \textit{(S)} is smoothing; \textit{(SS)} is symmetric smoothing; \textit{(ICP)} is scan-matching; \textit{(paper)} refers to the parameters as described in~\cite{adolfsson2023cfear3}; \textit{(github)} and \textit{(g)} refer to the parameters as in the Github repository of CFEAR-3; and (\dag) is the mean over all sequences except Sejong.}
\label{table:mulran_tr_deg}
\end{table*}

\begin{table*}[ht]
\begin{center}
\vspace{.3em} 
\def\arraystretch{1.1}
\begin{adjustbox}{width=1\textwidth}
\begin{tabular}{l|c c c | c c c  | c c c  | c c c |c c}
    & \multicolumn{2}{l}{\textit{Sequence}} & \multicolumn{1}{c}{} & & & \multicolumn{1}{c}{} & & & \multicolumn{1}{c}{} & & & \multicolumn{1}{c|}{} & &
    \\
    \hline
    \multirow{2}{*}{Method} & \multicolumn{3}{c|}{DCC} & \multicolumn{3}{c|}{KAIST} & \multicolumn{3}{c|}{Riverside} & \multicolumn{3}{c|}{Sejong} & \multirow{2}{*}{Mean$^{\dag}$} & \multirow{2}{*}{Mean}\\
     & 01 & 02 & \multicolumn{1}{c|}{03} & 01 & 02 & 03 & 01 & 02 & 03 & 01 & 02 & 03 & & \\
    \hline
    PhaRaO (full)~\cite{park2020pharao} & -/13.26 & -/13.26 & -/13.26 & -/12.8 & -/12.8 & -/12.8 & -/31.8 & -/31.8 & -/31.8 & - & - & - & -/19.3 & -\\
    CFEAR-1~\cite{adolfsson2021cfear1} & 8.22/21.64 & 5.00/15.83 & 5.77/20.77 & 7.53/37.06 & 7.22/23.28 & 7.47/38.79 & 7.98/30.81 & 7.36/49.25 & 6.39/129.23 & - & - & - & 6.99/40.74 & -\\
    CFEAR-2~\cite{adolfsson2021cfear2} & 7.99/17.32 & 4.70/7.95 & 5.46/6.81 & 7.18/17.26 & 6.70/16.58 & 6.96/23.80 & 7.39/22.06 & 6.86/47.28 & 5.83/80.52 & - & - & - & 6.56/26.62 & -\\
    CFEAR-3~\cite{adolfsson2023cfear3} & 7.83/6.82 & 4.54/5.13 & 5.16/4.88 & 6.37/8.72 & 6.01/9.89 & 6.21/13.44 & 5.90/10.98 & 5.38/3.26 & 4.50/17.83 & - & - & - & 5.77/8.99 & -\\
    \hline
    CFEAR-1*$_{paper}$ & 9.04/37.50 & 5.9/22.93 & 6.88/20.9 & 8.43/40.66 & 7.94/23.83 & 8.31/32.56 & 10.13/38.48 & 8.4/66.77 & 7.76/148.51 & 7.76/2741.56 & 6.6/2230.38 & 8.03/1863.47 & 8.09/48.02 & 7.93/605.63\\
    CFEAR-1*$_{github}$ & \textbf{8.09}/28.07 & \textbf{5.16}/14.89 & \textbf{6.12}/18.8 & \textbf{7.33}/34.99 & \textbf{7.0}/22.44 & 7.34/22.93 & 8.86/53.34 & 7.3/89.08 & 6.5/82.88 & \textbf{6.33}/3064.77 & \textbf{6.1}/2233.92 & \textbf{6.71}/1723.52 & \textbf{7.08}/40.83 & \textbf{6.9}/615.8\\ 
    CFEAR-2*$_{paper}$ & 8.77/24.40 & 5.7/12.05 & 6.61/10.37 & 8.0/18.1 & 7.43/23.26 & 7.85/19.67 & 9.58/20.97 & 8.09/45.55 & 7.16/76.84 & 8.37/2370.73 & 6.32/1769.7 & 6.99/1392.32 & 7.69/27.91 & 7.57/482.0\\
    CFEAR-2*$_{github}$ & 8.56/24.36 & 5.5/11.15 & 6.42/11.22 & 7.79/16.11 & 7.18/18.11 & 7.58/22.04 & 9.35/22.62 & 7.75/57.26 & 6.91/80.77 & 9.62/2285.38 & 6.19/2028.48 & 6.81/1470.76 & 7.45/29.29 & 7.47/504.02\\
    \hline
    CFEAR-3* & 8.73/6.56 & 5.53/\textbf{5.09} & 6.58/5.25 & 7.57/9.21 & 7.03/\textbf{11.5} & \textbf{7.32}/14.64 & \textbf{8.68}/6.86 & \textbf{7.25}/9.07 & \textbf{6.24}/19.54 & 6.38/854.37 & 6.18/737.08 & 6.77/129.12 & 7.21/9.75 & 7.02/150.69\\
    CFEAR-3*$_{S}$ & 11.29/10.46 & 7.83/7.27 & 9.38/7.68 & 9.71/12.29 & 9.37/16.2 & 9.6/15.76 & 11.39/10.83 & 10.03/18.26 & 9.6/9.82 & 7.94/727.12 & 7.75/602.56 & 8.27/\textbf{99.72} & 9.8/12.06 & 9.35/128.16\\
    CFEAR-3*$_{SS}$ & 8.79/6.75 & 5.68/5.11 & 6.65/5.34 & 7.65/10.41 & 7.15/11.77 & 7.39/\textbf{12.31} & 8.75/6.93 & 7.38/12.01 & 6.32/18.2 & 6.45/813.21 & 6.23/724.53 & 6.82/120.41 & 7.31/9.87 & 7.1/145.58\\
    CFEAR-3*$_{ICP}$ & 8.74/\textbf{6.55} & 5.97/5.12 & 6.62/\textbf{5.25} & 7.58/\textbf{9.21} & 7.03/11.5 & 7.32/14.64 & 8.69/\textbf{6.86} & 7.26/\textbf{9.06} & 6.38/19.0 & 6.48/856.67 & 6.23/737.03 & 6.82/129.38 & 7.29/\textbf{9.69} & 7.09/150.86\\
    CFEAR-3*$_{S + ICP}$ & 11.30/10.46 & 8.17/7.29 & 9.4/7.68 & 9.71/12.35 & 9.38/16.2 & 9.6/15.76 & 11.43/10.83 & 10.02/19.6 & 9.63/\textbf{8.77} & 8.03/\textbf{717.21} & 7.82/\textbf{602.32} & 8.3/101.36 & 9.85/12.11 & 9.4/\textbf{127.49}\\
    CFEAR-3*$_{SS + ICP}$ & 8.80/6.75 & 6.06/5.11 & 6.69/5.34 & 7.66/10.39 & 7.15/11.77 & 7.39/12.31 & 8.76/6.93 & 7.42/12.02 & 6.46/18.02 & 6.54/811.14 & 6.28/724.43 & 6.86/120.45 & 7.38/9.85 & 7.17/145.39\\
    \hline  
\end{tabular}
\end{adjustbox}
\end{center}
\caption{RPE [cm] and ATE [m] metrics, respectively, on the MulRan dataset~\cite{kim2020mulran}. \textit{(*)} is our re-implementation of CFEAR; \textit{(S)} is smoothing; \textit{(SS)} is symmetric smoothing; \textit{(ICP)} is scan-matching; \textit{(paper)} refers to the parameters as described in~\cite{adolfsson2023cfear3}; \textit{(github)} and \textit{(g)} refer to the parameters as described in the Github repository of CFEAR-3; and (\dag) is the mean over all sequences except Sejong.}
\label{table:mulran_rpe_ate}
\end{table*}

Contrary to Oxford, MulRan was only recorded using LiDAR and radar sensors, making it impossible for us to derive any qualitative insights about weather conditions as they are completely unknown. However, the most notable finding from this analysis, represented in Table~\ref{table:mulran_tr_deg} and Table~\ref{table:mulran_rpe_ate}, is the superior overall performance of ICP-based alternatives (e.g., ICP-only or ICP and kernel smoothing), emphasizing the importance of pose refinement in registration. This difference is particularly significant in the Sejong scene (which, to our knowledge, has never been used for evaluation in the literature), which features a large circular road with a diameter of 7 km. Here, the error accumulation is drastically reduced by the improvements we introduced in radar odometry compared.

An interesting thing to notice is that the lowest RPE is always achieved by the CFEAR-1* method, which uses the parameters from the corresponding GitHub repository (while for Oxford it was the CFEAR-2* algorithm).

\subsubsection{Boreas}

\begin{table*}[ht]
\begin{center}
\vspace{.3em} 
\def\arraystretch{1.1}
\begin{adjustbox}{width=1\textwidth}
\begin{tabular}{l|c c c c c c c c |c c}
    RPE [cm] / ATE [m]& \textit{Sequence} & & & & & & & & & \\
    \hline
    \multirow{2}{*}{Method} & Overcast, Snow & Overcast, Snow, Snowing & Overcast, Snow, Snowing & Sun & Sun, Clouds, Construction & Sun, Clouds & Night & Clouds & \multirow{2}{*}{Mean} & \multirow{2}{*}{Mean$^{\dag}$} \\
    & 11-26-13 & 12-01-13 & 01-26-11 & 04-08-12 & 04-13-14 & 06-03-16 & 09-14-20 & 11-16-14 & & \\
    \hline
    CFEAR-1*$_{paper}$ & 12.50/96.74 & 13.45/166.64 & 10.61/166.83 & 14.42/136.21 & 13.21/91.11 & 12.94/82.48 & 13.58/53.51 & 13.81/48.81 & 13.06/105.28 & 13.09/93.55\\
    CFEAR-1*$_{github}$ & \textbf{11.98}/131.85 & \textbf{12.89}/132.99 & \textbf{9.78}/187.25 & \textbf{13.88}/158.26 & \textbf{12.51}/92.76 & \textbf{12.40}/48.59 & \textbf{13.06}/48.25 & 13.48/\textbf{26.84} & \textbf{12.37}/103.35 & \textbf{12.53}/80.21\\
    CFEAR-2*$_{paper}$ & 12.41/55.24 & 13.32/53.34 & 10.29/94.23 & 14.35/36.94 & 12.91/47.80 & 12.86/56.75 & 13.56/110.99 & 13.74/73.41 & 12.93/66.15 & 12.95/64.91\\
    CFEAR-2*$_{github}$ & 12.10/49.28 & 12.94/97.41 & 9.89/113.25 & 13.98/75.04 & 12.58/43.17 & 12.49/44.76 & 13.11/39.50 & \textbf{13.44}/60.44 & 12.44/65.48 & 12.57/64.16\\
    \hline
    CFEAR-2*$_{g, S}$ & 13.82/48.47 & 14.56/32.35 & 11.84/32.11 & 15.36/24.35 & 14.01/46.04 & 13.86/57.32 & 14.54/130.11 & 14.28/113.09 & 13.78/60.48 & 14.01/82.11\\
    CFEAR-2*$_{g, SS}$ & 12.17/49.66 & 12.97/97.61 & 10.02/87.05 & 14.07/67.00 & 12.68/\textbf{29.20} & 12.52/\textbf{41.12} & 13.19/\textbf{34.13} & 13.47/63.03 & 12.64/\textbf{58.72} & 12.62/\textbf{60.72}\\
    CFEAR-2*$_{g, ICP}$ & 12.16/49.28 & 12.98/97.41 & 10.31/113.36 & 14.05/75.04 & 12.69/43.45 & 12.52/44.76 & 13.15/39.50 & 13.47/60.44 & 12.67/65.53 & 12.64/64.11\\
    CFEAR-2*$_{g, S + ICP}$ & 13.88/\textbf{48.44} & 14.61/\textbf{32.35} & 12.11/\textbf{31.41} & 15.45/\textbf{24.29} & 14.11/46.13 & 13.90/57.32 & 14.60/130.11 & 14.34/113.09 & 14.13/60.27 & 14.08/82.09\\
    CFEAR-2*$_{g, SS + ICP}$ & 12.22/49.66 & 13.01/97.62 & 10.46/89.39 & 14.13/67.00 & 12.80/29.21 & 12.55/41.12 & 13.23/34.13 & 13.51/63.03 & 12.74/58.77 & 12.70/60.78\\
    \hline
    CFEAR-3* & 14.79/107.52 & 15.80/114.38 & 13.34/91.32 & 16.82/79.05 & 15.68/94.98 & 14.77/135.09 & 15.60/106.15 & 13.48/99.98 & 15.16/103.44 & 14.93/110.86\\
    CFEAR-3*$_{S}$ & 16.06/119.07 & 17.07/120.68 & 15.07/92.58 & 17.89/98.07 & 16.85/82.45 & 15.94/125.92 & 16.76/53.89 & 14.16/94.10 & 16.35/98.35 & 16.07/103.62\\
    CFEAR-3*$_{SS}$ & 14.80/102.25 & 15.87/108.09 & 13.56/89.72 & 16.86/78.50 & 15.75/90.70 & 14.82/128.71 & 15.62/107.41 & 13.51/98.72 & 15.10/100.64 & 14.97/109.04\\
    CFEAR-3*$_{ICP}$ & 14.79/107.52 & 15.84/113.80 & 13.55/87.18 & 16.82/79.67 & 15.71/94.98 & 14.78/135.09 & 15.61/106.48 & 13.51/98.80 & 15.08/103.05 & 14.96/110.56\\
    CFEAR-3*$_{S + ICP}$ & 16.07/114.99 & 17.09/120.68 & 14.94/76.61 & 17.88/92.63 & 16.92/79.70 & 15.97/128.87 & 16.77/53.89 & 14.19/95.37 & 16.23/95.34 & 16.09/102.45\\
    CFEAR-3*$_{SS + ICP}$ & 14.81/102.25 & 15.90/108.09 & 13.71/97.93 & 16.88/78.44 & 15.77/90.70 & 14.83/128.71 & 15.65/107.60 & 13.53/97.68 & 15.13/101.43 & 15.00/109.28\\
    \hline
\end{tabular}
\end{adjustbox}
\end{center}
\caption{RPE [cm] and ATE [m] metrics, respectively, on the Boreas dataset~\cite{burnett_ijrr23}. \textit{(*)} is our re-implementation of CFEAR; \textit{(S)} is smoothing; \textit{(SS)} is symmetric smoothing; \textit{(ICP)} is scan-matching; \textit{(paper)} refers to the usage of parameters as described in~\cite{adolfsson2023cfear3}; \textit{(github)} and \textit{(g)} refer to the parameters as in the Github repository of CFEAR-3; and (\dag) is the mean over all sequences having ground truth.}
\label{table:boreas_rpe_ate}
\end{table*}

This dataset highlighted an interesting behavior of the existing algorithms (not studied before, because Boreas has never been used to evaluate methods for radar odometry estimation). CFEAR-3* has worse accuracy than CFEAR-1* and CFEAR-2*, as can be seen in Table~\ref{table:boreas_rpe_ate}. After an in-depth analysis, we discovered that the point-to-point metric used for registration and optimization leads to worse estimates w.r.t. the point-to-line metric used in CFEAR-1* and CFEAR-2* (other parameters do not seem to particularly influence the RPE/ATE). For this reason, we wanted to further stress the benefit of the proposed improvements by applying them to the CFEAR-2* algorithm that uses the parameters from the corresponding GitHub repository.

We utilized the extensive documentation of this dataset pertaining to weather conditions to derive qualitative insights. Notably, we were able to identify a distinct pattern in a chosen range of conditions. Specifically, the (S) + (ICP) configuration excels under poor lighting conditions (e.g., overcast), while in clear weather, the (SS) setting shows a marked improvement over other configurations. Moreover, from a quantitative standpoint, given that the majority of sequences were recorded under sunny conditions, we found that the (SS) configuration delivered the best performance.

In a similar way to what has been described for MulRan, also here the lowest RPE is always achieved by the CFEAR-1* method, with parameters from the GitHub repository.

Lastly, we decided to not include the translation percentage and orientation over distance errors, as they correspond to particularly high values (an average of 66.87 for the translation and 23.35 for the rotation). Nevertheless, the best accuracy is achieved by the (S) + (ICP) configuration and by CFEAR-1* (GitHub parameters), respectively.

\section{Conclusion}
In this paper, we presented a complete radar-only odometry estimation pipeline, built on CFEAR-3, and consisting of multiple steps: filtering, motion compensation, oriented surface points computation, smoothing, scan-to-multi-keyframe registration, and pose refinement. We developed an improved method for computing the set of oriented surface points, exploiting kernel-based smoothing techniques. We also proposed a pose refinement strategy that performs direct point cloud alignment between temporally consecutive radar scans. We evaluated our pipeline on all sequences of the Oxford Radar RobotCar and MulRan datasets, and on the majority of trajectories of the recent Boreas dataset, showing that we are able to achieve superior results in terms of accuracy in every scenario considered. We also performed an extensive qualitative and quantitative analysis, describing the impact and performance of our improvements under different environmental conditions. As a natural progression of this research, we plan to integrate the proposed pipeline into a SLAM framework in the near future and further improve its localization capabilities in complex environments.




\bibliographystyle{IEEEtran}
\bibliography{IEEEabrv, mybib}

\begin{thebibliography}{10}
\providecommand{\url}[1]{#1}
\csname url@rmstyle\endcsname
\providecommand{\newblock}{\relax}
\providecommand{\bibinfo}[2]{#2}
\providecommand\BIBentrySTDinterwordspacing{\spaceskip=0pt\relax}
\providecommand\BIBentryALTinterwordstretchfactor{4}
\providecommand\BIBentryALTinterwordspacing{\spaceskip=\fontdimen2\font plus
\BIBentryALTinterwordstretchfactor\fontdimen3\font minus \fontdimen4\font\relax}
\providecommand\BIBforeignlanguage[2]{{%
\expandafter\ifx\csname l@#1\endcsname\relax
\typeout{** WARNING: IEEEtran.bst: No hyphenation pattern has been}%
\typeout{** loaded for the language `#1'. Using the pattern for}%
\typeout{** the default language instead.}%
\else
\language=\csname l@#1\endcsname
\fi
#2}}

\bibitem{nahavandi2022comprehensive}
S.~Nahavandi, R.~Alizadehsani, D.~Nahavandi, S.~Mohamed, N.~Mohajer, M.~Rokonuzzaman, and I.~Hossain, ``A comprehensive review on autonomous navigation,'' \emph{arXiv preprint arXiv:2212.12808}, 2022.

\bibitem{garg2004detection}
K.~Garg and S.~K. Nayar, ``Detection and removal of rain from videos,'' in \emph{Proceedings of the 2004 IEEE Computer Society Conference on Computer Vision and Pattern Recognition, 2004. CVPR 2004.}, vol.~1.\hskip 1em plus 0.5em minus 0.4em\relax IEEE, 2004, pp. I--I.

\bibitem{charron2018noising}
N.~Charron, S.~Phillips, and S.~L. Waslander, ``De-noising of lidar point clouds corrupted by snowfall,'' in \emph{2018 15th Conference on Computer and Robot Vision (CRV)}.\hskip 1em plus 0.5em minus 0.4em\relax IEEE, 2018, pp. 254--261.

\bibitem{zhang2018robust}
C.~Zhang, M.~H. Ang, and D.~Rus, ``Robust lidar localization for autonomous driving in rain,'' in \emph{2018 IEEE/RSJ International Conference on Intelligent Robots and Systems (IROS)}.\hskip 1em plus 0.5em minus 0.4em\relax IEEE, 2018, pp. 3409--3415.

\bibitem{burnett2022we}
K.~Burnett, Y.~Wu, D.~J. Yoon, A.~P. Schoellig, and T.~D. Barfoot, ``Are we ready for radar to replace lidar in all-weather mapping and localization?'' \emph{IEEE Robotics and Automation Letters}, vol.~7, no.~4, pp. 10\,328--10\,335, 2022.

\bibitem{usuelli2023radarlcd}
M.~Usuelli, M.~Frosi, P.~Cudrano, S.~Mentasti, and M.~Matteucci, ``Radar{LCD}: Learnable radar-based loop closure detection pipeline,'' 2023.

\bibitem{venon2022millimeter}
A.~Venon, Y.~Dupuis, P.~Vasseur, and P.~Merriaux, ``Millimeter wave fmcw radars for perception, recognition and localization in automotive applications: A survey,'' \emph{IEEE Transactions on Intelligent Vehicles}, vol.~7, no.~3, pp. 533--555, 2022.

\bibitem{cen2018precise}
S.~H. Cen and P.~Newman, ``Precise ego-motion estimation with millimeter-wave radar under diverse and challenging conditions,'' in \emph{2018 IEEE International Conference on Robotics and Automation (ICRA)}.\hskip 1em plus 0.5em minus 0.4em\relax IEEE, 2018, pp. 6045--6052.

\bibitem{cen2019radar}
------, ``Radar-only ego-motion estimation in difficult settings via graph matching,'' in \emph{2019 International Conference on Robotics and Automation (ICRA)}.\hskip 1em plus 0.5em minus 0.4em\relax IEEE, 2019, pp. 298--304.

\bibitem{hong2022radarslam}
Z.~Hong, Y.~Petillot, A.~Wallace, and S.~Wang, ``Radarslam: A robust simultaneous localization and mapping system for all weather conditions,'' \emph{The international journal of robotics research}, vol.~41, no.~5, pp. 519--542, 2022.

\bibitem{adolfsson2023cfear3}
D.~Adolfsson, M.~Magnusson, A.~Alhashimi, A.~J. Lilienthal, and H.~Andreasson, ``Lidar-level localization with radar? the cfear approach to accurate, fast, and robust large-scale radar odometry in diverse environments,'' \emph{IEEE Transactions on Robotics}, vol.~39, no.~2, pp. 1476--1495, 2023.

\bibitem{barnes2020oxford}
D.~Barnes, M.~Gadd, P.~Murcutt, P.~Newman, and I.~Posner, ``The oxford radar robotcar dataset: A radar extension to the oxford robotcar dataset,'' in \emph{2020 IEEE International Conference on Robotics and Automation (ICRA)}.\hskip 1em plus 0.5em minus 0.4em\relax IEEE, 2020, pp. 6433--6438.

\bibitem{kim2020mulran}
G.~Kim, Y.~S. Park, Y.~Cho, J.~Jeong, and A.~Kim, ``Mulran: Multimodal range dataset for urban place recognition,'' in \emph{2020 IEEE International Conference on Robotics and Automation (ICRA)}.\hskip 1em plus 0.5em minus 0.4em\relax IEEE, 2020, pp. 6246--6253.

\bibitem{burnett_ijrr23}
K.~Burnett, D.~J. Yoon, Y.~Wu, A.~Z. Li, H.~Zhang, S.~Lu, J.~Qian, W.-K. Tseng, A.~Lambert, K.~Y. Leung, A.~P. Schoellig, and T.~D. Barfoot, ``Boreas: A multi-season autonomous driving dataset,'' \emph{The International Journal of Robotics Research}, vol.~42, no. 1-2, pp. 33--42, 2023.

\bibitem{caesar2020nuscenes}
H.~Caesar, V.~Bankiti, A.~H. Lang, S.~Vora, V.~E. Liong, Q.~Xu, A.~Krishnan, Y.~Pan, G.~Baldan, and O.~Beijbom, ``nuscenes: A multimodal dataset for autonomous driving,'' in \emph{Proceedings of the IEEE/CVF conference on computer vision and pattern recognition}, 2020, pp. 11\,621--11\,631.

\bibitem{kellner2013instantaneous}
D.~Kellner, M.~Barjenbruch, J.~Klappstein, J.~Dickmann, and K.~Dietmayer, ``Instantaneous ego-motion estimation using doppler radar,'' in \emph{16th International IEEE Conference on Intelligent Transportation Systems (ITSC 2013)}.\hskip 1em plus 0.5em minus 0.4em\relax IEEE, 2013, pp. 869--874.

\bibitem{jose2004relative}
E.~Jose and M.~D. Adams, ``Relative radar cross section based feature identification with millimeter wave radar for outdoor slam,'' in \emph{2004 IEEE/RSJ International Conference on Intelligent Robots and Systems (IROS)(IEEE Cat. No. 04CH37566)}, vol.~1.\hskip 1em plus 0.5em minus 0.4em\relax IEEE, 2004, pp. 425--430.

\bibitem{rouveure2009high}
R.~Rouveure, M.~Monod, and P.~Faure, ``High resolution mapping of the environment with a ground-based radar imager,'' in \emph{2009 International Radar Conference" Surveillance for a Safer World"(RADAR 2009)}.\hskip 1em plus 0.5em minus 0.4em\relax IEEE, 2009, pp. 1--6.

\bibitem{kung2021normal}
P.-C. Kung, C.-C. Wang, and W.-C. Lin, ``A normal distribution transform-based radar odometry designed for scanning and automotive radars,'' in \emph{2021 IEEE International Conference on Robotics and Automation (ICRA)}.\hskip 1em plus 0.5em minus 0.4em\relax IEEE, 2021, pp. 14\,417--14\,423.

\bibitem{aldera2019could}
R.~Aldera, D.~De~Martini, M.~Gadd, and P.~Newman, ``What could go wrong? introspective radar odometry in challenging environments,'' in \emph{2019 IEEE Intelligent Transportation Systems Conference (ITSC)}.\hskip 1em plus 0.5em minus 0.4em\relax IEEE, 2019, pp. 2835--2842.

\bibitem{park2020pharao}
Y.~S. Park, Y.-S. Shin, and A.~Kim, ``Pharao: Direct radar odometry using phase correlation,'' in \emph{2020 IEEE International Conference on Robotics and Automation (ICRA)}.\hskip 1em plus 0.5em minus 0.4em\relax IEEE, 2020, pp. 2617--2623.

\bibitem{burnett2021we}
K.~Burnett, A.~P. Schoellig, and T.~D. Barfoot, ``Do we need to compensate for motion distortion and doppler effects in spinning radar navigation?'' \emph{IEEE Robotics and Automation Letters}, vol.~6, no.~2, pp. 771--778, 2021.

\bibitem{adolfsson2021cfear1}
\BIBentryALTinterwordspacing
D.~Adolfsson, M.~Magnusson, A.~W. Alhashimi, A.~J. Lilienthal, and H.~Andreasson, ``Oriented surface points for efficient and accurate radar odometry,'' \emph{CoRR}, vol. abs/2109.09994, 2021. [Online]. Available: \url{https://arxiv.org/abs/2109.09994}
\BIBentrySTDinterwordspacing

\bibitem{adolfsson2021cfear2}
D.~Adolfsson, M.~Magnusson, A.~Alhashimi, A.~J. Lilienthal, and H.~Andreasson, ``Cfear radarodometry - conservative filtering for efficient and accurate radar odometry,'' in \emph{2021 IEEE/RSJ International Conference on Intelligent Robots and Systems (IROS)}, 2021, pp. 5462--5469.

\bibitem{alhashimi2021bfar}
A.~Alhashimi, D.~Adolfsson, M.~Magnusson, H.~Andreasson, and A.~J. Lilienthal, ``Bfar-bounded false alarm rate detector for improved radar odometry estimation,'' \emph{arXiv preprint arXiv:2109.09669}, 2021.

\bibitem{holder2019real}
M.~Holder, S.~Hellwig, and H.~Winner, ``Real-time pose graph slam based on radar,'' in \emph{2019 IEEE Intelligent Vehicles Symposium (IV)}.\hskip 1em plus 0.5em minus 0.4em\relax IEEE, 2019, pp. 1145--1151.

\bibitem{adolfsson2023tbv}
D.~Adolfsson, M.~Karlsson, V.~Kubelka, M.~Magnusson, and H.~Andreasson, ``Tbv radar slam – trust but verify loop candidates,'' \emph{IEEE Robotics and Automation Letters}, vol.~8, no.~6, pp. 3613--3620, 2023.

\bibitem{kim2018scan}
G.~Kim and A.~Kim, ``Scan context: Egocentric spatial descriptor for place recognition within 3d point cloud map,'' in \emph{2018 IEEE/RSJ International Conference on Intelligent Robots and Systems (IROS)}.\hskip 1em plus 0.5em minus 0.4em\relax IEEE, 2018, pp. 4802--4809.

\bibitem{arun1987least}
K.~S. Arun, T.~S. Huang, and S.~D. Blostein, ``Least-squares fitting of two 3-d point sets,'' \emph{IEEE Transactions on pattern analysis and machine intelligence}, no.~5, pp. 698--700, 1987.

\bibitem{aldera2022goes}
R.~Aldera, M.~Gadd, D.~De~Martini, and P.~Newman, ``What goes around: Leveraging a constant-curvature motion constraint in radar odometry,'' \emph{IEEE Robotics and Automation Letters}, vol.~7, no.~3, pp. 7865--7872, 2022.

\end{thebibliography}

\end{document}